\documentclass[preprint,12pt]{elsarticle}
\usepackage{amsmath,amssymb,bm,booktabs,array,graphicx,microtype}
\usepackage{enumitem}
\usepackage{pifont}
\usepackage[T1]{fontenc}
\setlist[itemize]{leftmargin=1.5em}

\journal{}
\begin{document}
\begin{frontmatter}
\title{ELVAE: Evidential Learning-Based Variational Autoencoder for Uncertainty-Aware Generation}
\author[rpi]{Ge Wang\corref{cor1}}
\ead{wangg6@rpi.edu}
\address[rpi]{Biomedical Imaging Center, Rensselaer Polytechnic Institute, 110 8th Street, Troy, 12180, New York, USA}
\cortext[cor1]{Corresponding author.}
\begin{abstract}
ELVAE places an input-dependent normal--inverse-gamma (NIG) hierarchy at each VAE latent coordinate, separating location uncertainty $u_{\mathrm{epi}}=\beta/[\nu(\alpha-1)]$ from conditional variability $u_{\mathrm{var}}=\beta/(\alpha-1)$. The marginalized latent law, however, identifies only the three quotient coordinates $(\gamma,\alpha,c)$ with $c=\beta(1+1/\nu)$; reconstruction is blind to one $(\nu,\beta)$ fiber direction. A companion theoretical analysis shows that the complete NIG prior and forward KL select a unique prior-relative canonical representative on each fiber, so canonical inverse allocation is not a fourth independent information channel. Empirically, trained inverse evidence $1/\nu$ remains the strongest sensitivity-ranking score. At $\tau_{\mathrm{epi}}=1$, the three-seed mean high/low-$u_{\mathrm{epi}}$ semantic-transition ratios are 1.98 on MNIST and 1.66 on Fashion-MNIST, falling to 1.33 and 1.16 under scale-matched controls. In the 20-draw MNIST component study with equal per-anchor perturbation energy, $1/\nu$ gives high/low ratios 1.69 and 1.65 under the $u_{\mathrm{epi}}$ and $u_{\mathrm{var}}$ fields and 1.69 (95\% interval 1.45--1.97) under a geometry-free isotropic field, whereas $u_{\mathrm{var}}$ reverses the isotropic ordering to 0.76. Under the experimental prior, canonical $1/\nu_{\mathrm{can}}$ is a strictly increasing transform of $T=c/[\alpha(\gamma^2+2)]$ and is bounded above by $3+\sqrt{10}$. Thus ELVAE exposes a controllable sensitivity mechanism whose trained four-output realization is operationally informative, while the exact reconstruction-visible information remains three-dimensional and baseline image quality is a separate question.
\end{abstract}
\end{frontmatter}

\section{Introduction}

Variational autoencoders (VAEs) represent an observation $y$ using a probabilistic latent variable $z$ and generate samples by decoding latent draws [1, 2]. A standard diagonal-Gaussian encoder has the form

\begin{equation}
q_\phi(z\mid y)=\mathcal N\!\left(\mu_\phi(y),\operatorname{diag}\sigma_\phi^2(y)\right). \tag{1}
\end{equation}

The posterior variance in Eq. (1) supplies stochasticity, but it does not separately represent uncertainty about where the latent center should be and variability around a given center. This distinction matters for posterior-anchored generation. Two anchors may admit similar total latent spread while differing in how well their latent locations are determined. Evidential learning represents higher-order uncertainty by predicting parameters of a distribution over lower-order probabilistic quantities [3, 4]. Posterior and Natural Posterior Networks develop related single-pass evidential constructions, while Ulmer et al. survey this broader family [5, 6, 7]. Normal-inverse-gamma models are particularly attractive for continuous variables because they yield an analytic decomposition between variance of the mean and expected conditional variance. Nevertheless, evidential uncertainty is not automatically trustworthy. Loss-minimization pathologies and non-identifiability can make apparently interpretable parameters weakly related to empirical error [8, 9]. Evaluation must therefore separate the mathematical existence of an uncertainty quantity from its operational value. This paper develops and tests ELVAE, an evidential learning-based VAE in which every latent coordinate is controlled by an input-dependent NIG posterior. The method is related to evidential latent-variable models [10, 11] and, through its marginal, to heavy-tailed VAEs. Marginalizing $(\mu,\sigma^2)$ makes the latent law Student-$t$ with $2\alpha(y)$ degrees of freedom, and the prior NIG(0, 1, 3, 1) marginalizes to a fixed Student-$t$ with six degrees of freedom and unit variance; ELVAE therefore does not claim a new latent distribution class [12, 13, 14, 15]. The contributions are narrower. First, the input-dependent, coordinate-wise NIG hierarchy supplies separate latent-location and conditional-variability terms that can be used as controlled perturbation fields. Second, the reconstruction-visible family is only the three-coordinate quotient $(\gamma,\alpha,c)$; the companion theoretical paper establishes the unique prior-relative KL section, exact three-coordinate reduction, monotone quotient score, and residual prior gauge [21]. The present paper uses those results rather than reproducing their proofs. Third, the trained ratio $u_{\mathrm{epi},k}/u_{\mathrm{var},k}=1/\nu_k$ is the strongest empirical sensitivity-ranking variable, creating a useful distinction between the exact canonical geometry and the finite four-output amortized realization. Fourth, the evaluation is designed around posterior-anchored generation and explicitly distinguishes five questions:

\begin{enumerate}
\item Does $u_{\mathrm{epi}}$ rank the reliability of the zero-displacement anchor generation $g_\theta(\gamma)$?
\item Does $u_{\mathrm{epi}}$ stratify sensitivity when it is used to scale controlled latent perturbations?
\item How much of that stratification remains when perturbation amplitude is fixed or permuted, breaking the direct link between score and displacement magnitude?
\item Do $u_{\mathrm{epi}}$ and $u_{\mathrm{var}}$ behave differently when each is perturbed at matched latent energy, and which component ranks anchor sensitivity?
\item Does the stratification survive when all learned coordinate geometry is removed from the perturbation field, so that the ranking score cannot act through the field it scales?
\end{enumerate}

\subsection{Relation to heavy-tailed VAEs}

Student-$t$ distributions have entered VAEs through several distinct routes. Takahashi et al. [12] use a Student-$t$ observation model to improve robustness in density estimation. Abiri and Ohlsson [13] introduce a trainable multivariate Student-$t$ prior, while Mathieu et al. [14] use products of Student-$t$ priors to impose structured latent decompositions. The $t^3$ VAE of Kim et al. [15] places Student-$t$ distributions in the prior, encoder, and decoder and replaces KL-based training with a power-divergence objective. ELVAE does not claim a new marginal distribution. Its distinction is the input-dependent, coordinate-wise NIG hierarchy; direct regularization of that higher-order posterior; and identification of the resulting inverse-evidence allocation $1/\nu$ as the variable that governs anchor sensitivity under energy-matched and geometry-free perturbation. The first version of the study addressed these questions only on MNIST and used one perturbation draw per anchor. The present work substantially expands the evidence. It adds Fashion-MNIST as a distinct object-image dataset [16], increases training to stable convergence, uses five draws per anchor at five perturbation amplitudes, reports both correct-to-wrong degradation and wrong-to-correct correction, compares uncertainty proxies and a conventional VAE, adds fixed-scale and permutation controls, and repeats the complete ELVAE pipeline for three seeds on both datasets. A focused MNIST experiment further uses 20 paired draws per anchor to compare both NIG variance components under population-level and per-anchor energy matching, identifies inverse evidence $1/\nu$ as the stronger sensitivity-ranking mechanism, and checks the result with a high-accuracy CNN classifier. The resulting conclusion is more specific than the usual claim that ``high uncertainty means bad generation.'' The two-dataset evidence supports $u_{\mathrm{epi}}$ as a population-level perturbation-sensitivity variable. It does not support $u_{\mathrm{epi}}$ as a universal anchor-quality score. This distinction is important for using uncertainty in augmentation and stress testing: a useful sampling control need not be the best detector of baseline reconstruction failure.

\section{Evidential Latent Hierarchy}

\subsection{Normal-inverse-gamma encoder}

For each latent coordinate $k=1,\ldots,K$, the encoder predicts

\begin{equation}
(\gamma_k,\nu_k,\alpha_k,\beta_k),\qquad \nu_k>0,\quad \alpha_k>1,\quad \beta_k>0. \tag{2}
\end{equation}

which parameterize

\begin{align}
\sigma_k^2\mid y &\sim \operatorname{InvGamma}(\alpha_k,\beta_k),\nonumber\\
\mu_k\mid\sigma_k^2,y &\sim \mathcal N\!\left(\gamma_k,\frac{\sigma_k^2}{\nu_k}\right),\nonumber\\
z_k\mid\mu_k,\sigma_k^2 &\sim \mathcal N(\mu_k,\sigma_k^2). \tag{3}
\end{align}

Thus $q_\phi(\mu_k,\sigma_k^2\mid y)=\operatorname{NIG}(\gamma_k,\nu_k,\alpha_k,\beta_k)$. Positivity is enforced with softplus transforms, with a +1 offset on $\alpha$ so that the relevant expectations exist. The hierarchy defines

\begin{align}
u_{\mathrm{var},k} &\equiv \mathbb E[\sigma_k^2\mid y]=\frac{\beta_k}{\alpha_k-1}, \tag{4}\\
u_{\mathrm{epi},k} &\equiv \operatorname{Var}(\mu_k\mid y)=\frac{\beta_k}{\nu_k(\alpha_k-1)}, \tag{5}\\
\operatorname{Var}(z_k\mid y) &= u_{\mathrm{var},k}+u_{\mathrm{epi},k}. \tag{6}
\end{align}

The two components also define the coordinate-wise inverse-evidence ratio

\begin{equation}
r_k(y)\equiv\frac{u_{\mathrm{epi},k}(y)}{u_{\mathrm{var},k}(y)}=\frac{1}{\nu_k(y)}. \tag{7}
\end{equation}

Unlike either variance component alone, $r_k$ cancels the shared scale $\beta_k/(\alpha_k-1)$ and isolates how strongly the hierarchy allocates uncertainty to the latent location rather than conditional spread. We use $K^{-1}\sum_k r_k$ as the scalar inverse-evidence score. The identity in Eq. (7) is coordinate-wise: because averaging and division do not commute, scalar $u_{\mathrm{epi}}(y)$ is not identically proportional to mean inverse evidence. Their empirical correlation is 0.925 in the detailed MNIST run. We call Eq. (5) epistemic latent uncertainty because it measures uncertainty in the latent location itself. For ranking one image, we use the coordinate mean

\begin{equation}
u_{\mathrm{epi}}(y)=\frac{1}{K}\sum_{k=1}^{K}u_{\mathrm{epi},k}(y). \tag{8}
\end{equation}

\subsection{Training objective and ELBO interpretation}

With $P$ input pixels, the training objective is

\begin{equation}
\mathcal L_{\mathrm{ELVAE}}(y)=\mathcal L_{\mathrm{recon}}(y)+\lambda_{\mathrm{NIG}}\mathcal L_{\mathrm{NIG}}(y). \tag{9}
\end{equation}
where
\begin{equation}
\mathcal L_{\mathrm{recon}}=\frac{1}{P}\mathbb E\|y-g_\theta(z)\|_2^2, \tag{10}
\end{equation}
\begin{equation}
\mathcal L_{\mathrm{NIG}}=\frac{1}{K}\sum_{k=1}^{K}D_{\mathrm{KL}}\!\left[q_\phi(\mu_k,\sigma_k^2\mid y)\,\|\,p_0(\mu_k,\sigma_k^2)\right]. \tag{11}
\end{equation}

We use the fixed prior $(\gamma_0,\nu_0,\alpha_0,\beta_0)=(0,1,3,1)$, for which $\mathbb E[\sigma^2]=1/2$, $\operatorname{Var}(\mu)=1/2$, and $\operatorname{Var}(z)=1$. Equation (9) is an exact ELBO objective for the hierarchical generative model

\begin{equation}
p(\mu,\sigma^2)=p_0,\qquad p(z\mid\mu,\sigma^2)=\mathcal N(\mu,\sigma^2),\qquad p(y\mid z)=\mathcal N(g_\theta(z),s^2I_P). \tag{12}
\end{equation}

Choosing the inference conditional $q(z\mid\mu,\sigma^2,y)=p(z\mid\mu,\sigma^2)$ causes the conditional KL to cancel. Up to constants, the negative ELBO is

\begin{equation}
-\mathrm{ELBO}(y)=\frac{P}{2s^2}\mathcal L_{\mathrm{recon}}+K\mathcal L_{\mathrm{NIG}}+\mathrm{constant}. \tag{13}
\end{equation}

Matching coefficients yields

\begin{equation}
\lambda_{\mathrm{NIG}}=\frac{2s^2K}{P}. \tag{14}
\end{equation}

The experiments use $\lambda_{\mathrm{NIG}}=5\times10^{-4}$. By Eq. (14), this corresponds to a fixed decoder noise variance $s^2=\lambda_{\mathrm{NIG}}P/(2K)$: $2.45\times10^{-2}$ for MNIST ($K=8$) and $1.23\times10^{-2}$ for Fashion-MNIST ($K=16$). Thus a common $\lambda_{\mathrm{NIG}}$ fixes the optimization weight across runs but implies different decoder variances when $K$ differs. In this ELBO interpretation, $s^2$ is a fixed model hyperparameter; it is not estimated from the deterministic reconstruction $g_\theta(\gamma)$, and no equality with its empirical residual is assumed. The controlled Gaussian perturbations introduced in Section 3 are evaluation constructions, not a replacement for the probabilistic training model in Eq. (3). The ELBO interpretation above is defined by the hierarchical latent law; the variance-matched fields below are used only to isolate operational sensitivity under controlled displacement. Training uses the hierarchical reparameterization in Eq. (3): precision $\lambda=1/\sigma^2$ is drawn from $\operatorname{Gamma}(\alpha,\beta)$, followed by $\mu=\gamma+\sqrt{\sigma^2/\nu}\,\epsilon_1$ and $z=\mu+\sqrt{\sigma^2}\,\epsilon_2$, with independent standard-normal $\epsilon_1,\epsilon_2$. The controlled Gaussian fields in Section 3 are used only for evaluation.

\subsection{Quotient non-identifiability and the role of hierarchical regularization}

Marginalizing $(\mu,\sigma^2)$ gives a Student-$t$ distribution for $z$ with $2\alpha$ degrees of freedom, location $\gamma$, and squared scale $\beta(1+1/\nu)/\alpha$. The exact marginal therefore depends on the four NIG parameters only through

\begin{equation}
(\gamma,\alpha,c),\qquad c=\beta\left(1+\frac{1}{\nu}\right). \tag{15}
\end{equation}

For fixed $(\gamma,\alpha,c)$, the curve $\beta=c\nu/(1+\nu)$ is a one-dimensional fiber along which the marginal distribution of $z$ and every exact reconstruction functional are unchanged. Along this fiber,

\begin{equation}
u_{\mathrm{var}}=\frac{c}{\alpha-1}\frac{\nu}{1+\nu},\qquad
u_{\mathrm{epi}}=\frac{c}{\alpha-1}\frac{1}{1+\nu}. \tag{16}
\end{equation}

can trade continuously. A loss defined only on the Student-$t$ marginal therefore identifies three reconstruction-visible coordinates, not four independent hierarchy coordinates. The full NIG-to-NIG KL in Eq. (9) supplies additional prior-relative structure. For a fixed complete hierarchical prior, minimizing that same forward KL along each fiber selects a unique representative, derived in Section 2.4 and in the companion theoretical study [21]. This is more precise than saying that regularization simply ``breaks'' the invariance: the quotient is intrinsic to the marginal law, whereas the selected section is canonical only relative to the specified prior and variational objective. Exact expected likelihood is fiber-invariant; finite-sample or reparameterized gradient estimators may nevertheless have split-dependent variance and finite amortized networks need not sit exactly on the analytic section. This distinction also fixes how $\nu_k$ should be interpreted. In a conjugate NIG update, $\nu_k$ has the algebraic role of an effective pseudo-observation count controlling the precision of the latent location. In ELVAE it is amortized directly by the encoder rather than accumulated through Bayesian updating. At the level of the exact variational family, the canonical $\nu_{\mathrm{can}}$ is determined by $(\gamma,\alpha,c)$ and the complete prior; it is not a fourth independent information channel supplied by the reconstruction likelihood. The trained four-output network can deviate from this section because of finite parameterization and optimization. We therefore use $1/\nu_k$ operationally as a trained sensitivity score, while withholding any guarantee that it is calibrated to reconstruction quality.

\subsection{Canonical allocation and quotient score}

The companion theoretical study gives the general-prior derivation, uniqueness proof, exact partial minimization, and residual-gauge analysis [21]. Here we record only the specialization needed to interpret the experiments. For the fixed prior $p_0=\operatorname{NIG}(0,1,3,1)$ and a quotient state $(\gamma,\alpha,c)$, define

\begin{equation}
B=\frac12+\frac{\alpha}{c}\left(1+\frac{\gamma^2}{2}\right). \tag{17}
\end{equation}

The forward KL restricted to the fiber $\beta=c\nu/(1+\nu)$ has the unique global minimizer

\begin{equation}
\nu_{\mathrm{can}}=B-\frac72+\sqrt{\left(B-\frac72\right)^2+2B}, \tag{18}
\end{equation}
\begin{equation}
\beta_{\mathrm{can}}=c\frac{\nu_{\mathrm{can}}}{1+\nu_{\mathrm{can}}}. \tag{19}
\end{equation}

Partial minimization over this fiber is exact at the level of the underlying variational family: replacing the four-coordinate objective by its fiber-minimized three-coordinate form preserves the optimum for every positive KL weight [21]. This does not imply identical optimization behavior for arbitrary finite three-output and four-output neural networks. For the same prior, define

\begin{equation}
T=\frac{c}{\alpha(\gamma^2+2)}. \tag{20}
\end{equation}
Then $B=\tfrac12(1+1/T)$ and
\begin{equation}
\frac{1}{\nu_{\mathrm{can}}}=g(T) \tag{21}
\end{equation}

is strictly increasing in $T$ [21]. Thus canonical inverse allocation is a prior-calibrated re-expression of the quotient state rather than a fourth independent information coordinate. The same theory gives the prior-specific amplitude bound

\begin{equation}
0<\frac{1}{\nu_{\mathrm{can}}}<3+\sqrt{10}\approx6.162. \tag{22}
\end{equation}

Any trained coordinate above this ceiling is therefore necessarily off the canonical section, although remaining below the ceiling alone does not establish proximity to that section. This ceiling provides a direct falsification audit for the archived trained allocation: because $1/\nu_k=u_{\mathrm{epi},k}/u_{\mathrm{var},k}$, no encoder pass is needed for the bound check itself.

\section{Controlled Uncertainty-Aware Generation}

For exact posterior generation, one samples $\sigma^2$ , then $\mu$, then $z$ according to Eq. (3). The experiments below deliberately do not use that exact draw when testing sensitivity. For controlled evaluation, we instead use a variance-matched perturbation centered at the encoder location,

\begin{equation}
z=\gamma+\tau_{\mathrm{epi}}\sqrt{u_{\mathrm{epi}}}\odot\epsilon_{\mathrm{epi}},\qquad \epsilon_{\mathrm{epi}}\sim\mathcal N(0,I). \tag{23}
\end{equation}

This is not an exact Student-$t$ draw. It is a control parameterization whose coordinate-wise perturbation variance equals $\tau_{\mathrm{epi}}^2 u_{\mathrm{epi},k}$. A more general two-component form is

\begin{equation}
z=\gamma+\tau_{\mathrm{epi}}\sqrt{u_{\mathrm{epi}}}\odot\epsilon_{\mathrm{epi}}+\tau_{\mathrm{var}}\sqrt{u_{\mathrm{var}}}\odot\epsilon_{\mathrm{var}}. \tag{24}
\end{equation}

The primary two-dataset experiments set $\tau_{\mathrm{var}}=0$ to isolate the latent-location component. A focused MNIST component experiment separately activates $u_{\mathrm{epi}}$ or $u_{\mathrm{var}}$. Let $u_{c,ik}$ denote component $c\in\{\mathrm{epi},\mathrm{var}\}$ for anchor $i$ and coordinate $k$, and let

\begin{equation}
E_{\mathrm{epi}}=\frac1N\sum_{i=1}^{N}\sum_{k=1}^{K}u_{\mathrm{epi},ik}. \tag{25}
\end{equation}

For population-level energy matching, each component is rescaled as

\begin{equation}
\widetilde u_{c,ik}=u_{c,ik}\frac{E_{\mathrm{epi}}}{N^{-1}\sum_{j=1}^{N}\sum_{\ell=1}^{K}u_{c,j\ell}}. \tag{26}
\end{equation}

For the stricter per-anchor control,

\begin{equation}
\widehat u_{c,ik}=u_{c,ik}\frac{E_{\mathrm{epi}}}{\sum_{\ell=1}^{K}u_{c,i\ell}}, \tag{27}
\end{equation}

so every anchor and either component satisfy $\mathbb E\|\Delta z_i\|_2^2=\tau^2E_{\mathrm{epi}}$. Equation (27) removes component- and anchor-level amplitude differences while retaining coordinate-wise allocation. Common Gaussian noise is used for paired $u_{\mathrm{epi}}/u_{\mathrm{var}}$ comparisons. To remove learned coordinate allocation as well, we add an isotropic field

\begin{equation}
\widehat u_{\mathrm{iso},ik}=\frac{E_{\mathrm{epi}}}{K}, \tag{28}
\end{equation}

which has the same expected energy for every anchor but contains no learned variance geometry. Persistence of score stratification under Eq. (28) therefore identifies an anchor-level association rather than a coupling between the ranking score and perturbation field. The same $u_{\mathrm{epi}}$ has two roles in Eq. (23): it ranks anchors and scales their perturbations. We therefore distinguish:

\begin{itemize}
\item \textbf{Zero-displacement control (C):} $z=\gamma$. The decoder still generates $g_\theta(\gamma)$ and $u_{\mathrm{epi}}$ is retained for ranking.
\item \textbf{Uncertainty-scaled generation (A):} Eq. (23) at a specified $\tau_{\mathrm{epi}}$.
\item \textbf{Induced failure (I):} condition (A) is wrong among anchors for which condition (C) is correct.
\item \textbf{Induced correction (R):} condition (A) is correct among anchors for which condition (C) is wrong.
\item \textbf{Semantic transition (T):} the frozen classifier label under (A) differs from its label under (C), regardless of whether either label equals the intended class.
\end{itemize}

The transition statistic is bidirectional and avoids treating every perturbation as damage. It includes correct-to-wrong, wrong-to-correct, and wrong-class-to-different-wrong-class changes. To test whether high/low separation is merely caused by high-$u_{\mathrm{epi}}$ anchors receiving larger displacements, we add two controls at $\tau_{\mathrm{epi}}=1$. The fixed-class-median control gives every anchor in a class the same coordinate-wise median $u_{\mathrm{epi}}$ scale. The within-class-permuted control randomly reassigns complete $u_{\mathrm{epi}}$ vectors among anchors of the same class. Both preserve class structure while breaking the direct score--amplitude assignment.

\section{Experimental Design}

\subsection{Datasets and models}

MNIST contains 60,000 training and 10,000 test images of handwritten digits [17]. Fashion-MNIST uses the same split and image size but contains ten apparel classes [16]. ELVAE training is unsupervised; class labels are used only for stratified evaluation and the external classifier. The MNIST ELVAE uses an MLP encoder 784 $\to$ 128 $\to$ 64, latent dimension $K$ = 8, and a mirrored decoder. It is trained for eight epochs with Adam, learning rate $10^{-3}$, batch size 1024, and $\lambda_{\mathrm{NIG}}=5\times10^{-4}$. The Fashion-MNIST model uses the same hidden layers, $K$ = 16, twelve epochs, and batch size 512. A matched Gaussian VAE uses the same encoder and decoder widths and the same regularization weight. Separate classifiers 784 $\to$ 256 $\to$ 128 $\to$ 10 are trained only on real training images and then frozen. Their held-out accuracies are 94.79\% for MNIST and 86.55\% for Fashion-MNIST. The MNIST robustness classifier has three convolutional stages and two fully connected layers and attains 98.84\% held-out accuracy. For each held-out anchor, five independent perturbations are generated at $\tau_{\mathrm{epi}}\in\{0,0.5,1,1.5,2\}$. The scaled runs and the scale-matched controls of Section 5.3 draw their noise from separate fixed generators, so the realizations of $\epsilon$ differ between them; with 10,000 anchors and five draws this contributes negligibly to the reported rates. Every result is computed from the same 10,000 anchors. The scalar $u_{\mathrm{epi}}$ in Eq. (8) is ranked within intended class to prevent class-level uncertainty offsets from confounding the pooled groups. The complete ELVAE training and evaluation are repeated for seeds 1, 2, and 20260809. The frozen classifier is shared across ELVAE seeds within a dataset. The component experiment reuses the detailed MNIST checkpoint (seed 20260809), all 10,000 anchors, and the same frozen classifier. It evaluates 20 paired perturbations per anchor at $\tau\in\{0.5,1,1.5,2\}$. High/low groups are formed within intended class using scalar $u_{\mathrm{epi}}$, $u_{\mathrm{var}}$, or mean inverse evidence $K^{-1}\sum_k 1/\nu_k$. The $u_{\mathrm{epi}}$, $u_{\mathrm{var}}$, and isotropic fields share both expected energy and common Gaussian draws. Ratios receive 95\% percentile intervals from 3,000 anchor-level bootstrap replicates; resampling anchors preserves dependence among the 20 draws from one anchor. As a semantic-proxy robustness check, the anchor-energy-matched $\tau=1$ comparison is repeated with a separately trained convolutional classifier using the same anchors, learned fields, and five common noise draws. Because the MLP and CNN analyses use 20 and five draws, respectively, the CNN check tests ordering direction only; their effect sizes are not directly compared.

\subsection{Compact canonical-closure ablation}

The original detailed MNIST arrays retained $u_{\mathrm{epi},k}$, $u_{\mathrm{var},k}$, and $\gamma_k$ but not $\alpha_k$ and $c_k$ separately, so they cannot reconstruct the exact canonical score after the fact. We therefore add a small independent closure experiment in which all four quantities are saved explicitly. The scikit-learn handwritten-digits dataset contains 1,797 $8\times8$ grayscale digit images; a stratified split uses 1,347 for training and 450 for testing [22]. Both models use two 128-unit hidden layers and latent dimension $K$ = 8. The four-output model predicts $(\gamma,\nu,\alpha,\beta)$; the canonical three-output model predicts $(\gamma,\alpha,c)$ and inserts Eqs. (17)--(18) and $\beta_{\mathrm{can}}=c\nu_{\mathrm{can}}/(1+\nu_{\mathrm{can}})$ before sampling and KL evaluation. Both are trained for 120 epochs with Adam, learning rate $10^{-3}$, batch size 128, $\lambda_{\mathrm{NIG}}=0.02$, and seed 20260809. A frozen SVM classifier trained only on the real training images reaches 98.44\% held-out accuracy and is used only as a semantic-transition proxy. Twenty common Gaussian perturbation draws at $\tau=1$ are evaluated with the same per-anchor energy matching as Eqs. (27)--(28). This compact experiment is a targeted architecture and geometry check; it does not replace the larger MNIST/Fashion-MNIST study or its reported effect sizes.

\subsection{Metrics and baselines}

We report classifier error at $z=\gamma$, low/high 20\% ratios, AUROC, semantic-transition rate, induced-failure rate, induced-correction rate, and dose response across $\tau_{\mathrm{epi}}$. Baseline anchor scores are reconstruction error, $u_{\mathrm{var}}$, total latent variance $u_{\mathrm{epi}}+u_{\mathrm{var}}$, and $\|\gamma\|$. A matched Gaussian VAE supplies posterior variance, reconstruction error, and latent norm baselines for zero-displacement anchor reliability. Detailed-run ratios and the paired component study receive 95\% anchor-bootstrap intervals; the complete-seed table reports the observed range across independently trained models.

\section{Results}

\subsection{Qualitative examples}

Figures 1 and 2 show anchors selected near the 10th and 90th within-class $u_{\mathrm{epi}}$ percentiles, without selecting on classifier outcome. Each panel displays the original anchor, zero-displacement generation $g_\theta(\gamma)$, and one $\tau_{\mathrm{epi}}=1$ generation. Both models produce visibly smoothed outputs, as expected from compact pixel-MSE VAEs. The figures also illustrate why semantic transitions, rather than subjective sharpness alone, are used for the main comparison.

\begin{figure}[p]
\centering\includegraphics[width=0.96\linewidth]{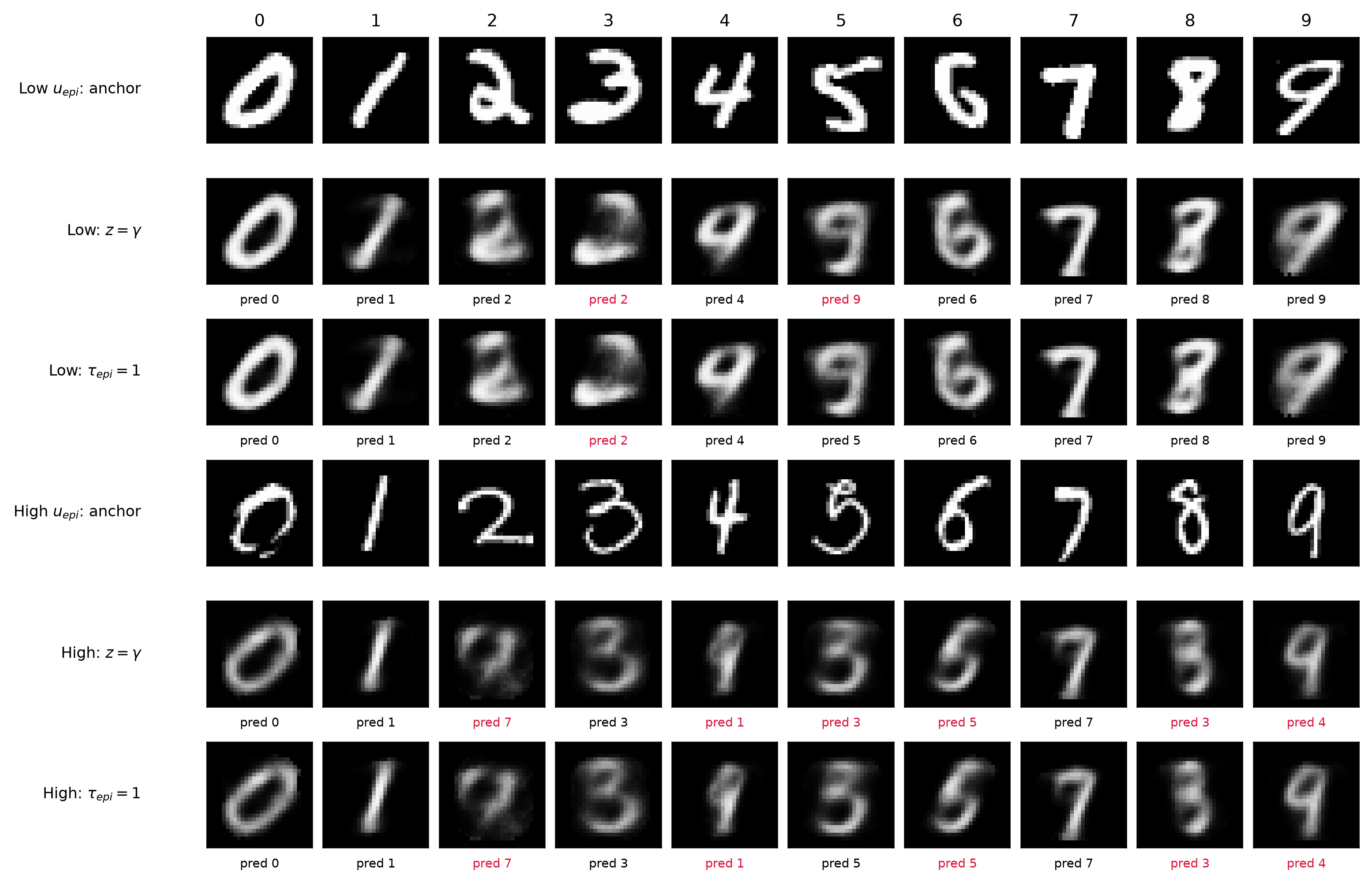}
\caption{MNIST examples selected near the 10th (low) and 90th (high) within-class $u_{\mathrm{epi}}$ percentiles. Rows show the original anchor, zero-displacement generation $g_\theta(\gamma)$, and one uncertainty-scaled generation at $\tau_{\mathrm{epi}}=1$. Red predictions differ from the intended class. Selection is based only on uncertainty percentile.}
\label{fig:mnist}
\end{figure}
\begin{figure}[p]
\centering\includegraphics[width=0.96\linewidth]{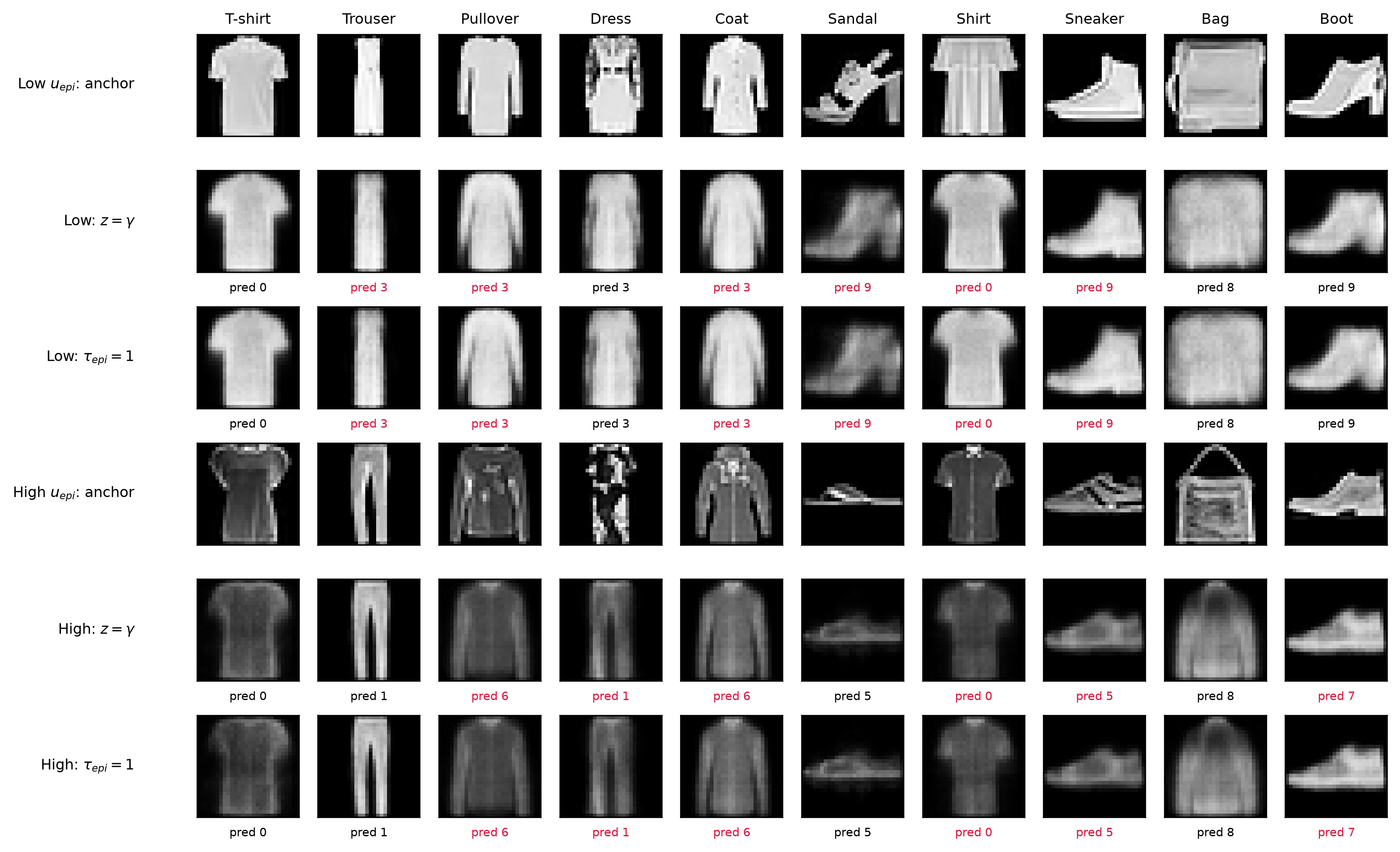}
\caption{Fashion-MNIST examples under the same selection and generation protocol as Fig. 1. High-$u_{\mathrm{epi}}$ anchors are not uniformly worse at $z=\gamma$; the quantitative result concerns their greater sensitivity to controlled perturbation.}
\label{fig:fashion}
\end{figure}
\clearpage

\subsection{Primary perturbation-sensitivity result}

Table 1 reports the detailed seed at $\tau_{\mathrm{epi}}=1$. On MNIST, the classifier-label transition rate is 3.62\% in the low-$u_{\mathrm{epi}}$ group and 6.58\% in the high-$u_{\mathrm{epi}}$ group, a ratio of $1.82\times$. On Fashion-MNIST the corresponding rates are 3.08\% and 5.61\%, also $1.82\times$. Restricting to anchors correctly generated at $z$ = $\gamma$, induced-failure ratios are $1.78\times$ and $1.89\times$. Thus the cross-dataset agreement is strongest for perturbation sensitivity, not for baseline anchor quality.

\begin{table}[t]
\centering\small
\caption{Detailed-run results at $\tau_{\mathrm{epi}}=1$; each percentage averages five draws per anchor. Brackets give 95\% anchor-bootstrap intervals for the high/low ratio.}
\resizebox{\linewidth}{!}{%
\begin{tabular}{llrrrr}
\toprule
Dataset & Quantity & Low 20\% & High 20\% & Ratio & 95\% interval\\
\midrule
MNIST & Semantic transition & 3.62\% & 6.58\% & 1.82 & [1.52, 2.20]\\
& Induced failure ($C\checkmark\to A\times$) & 2.05\% & 3.65\% & 1.78 & [1.35, 2.40]\\
& Induced correction ($C\times\to A\checkmark$) & 6.25\% & 7.88\% & 1.26 & [0.95, 1.72]\\
& Total error under A & 22.59\% & 29.13\% & 1.29 & [1.17, 1.43]\\
\addlinespace
Fashion-MNIST & Semantic transition & 3.08\% & 5.61\% & 1.82 & [1.50, 2.23]\\
& Induced failure ($C\checkmark\to A\times$) & 1.75\% & 3.31\% & 1.89 & [1.40, 2.69]\\
& Induced correction ($C\times\to A\checkmark$) & 3.79\% & 8.16\% & 2.15 & [1.58, 3.01]\\
& Total error under A & 30.18\% & 24.91\% & 0.83 & [0.75, 0.91]\\
\bottomrule
\end{tabular}%
}
\label{tab:primary}
\end{table}

Correction events are not negligible in absolute frequency, confirming that zero-mean latent perturbation does not simply ``make images worse.'' The MNIST high/low correction ratio is 1.26, but its 95\% interval [0.95, 1.72] includes one; only Fashion-MNIST shows a clear group difference (2.15, [1.58, 3.01]). The total error nevertheless increases slightly from C to A in both datasets: 0.41 percentage points on MNIST and 0.26 points on Fashion-MNIST across all anchors. The net changes are small because degradation and correction partly cancel. Figure 3 gives the dose response averaged across three seeds, with the shaded region spanning the minimum and maximum seed. Transition rates rise with $\tau_{\mathrm{epi}}$ in both uncertainty groups, but the high-$u_{\mathrm{epi}}$ slope is consistently steeper. At $\tau_{\mathrm{epi}}=2$, the mean transition rates reach approximately 14.9\% versus 7.4\% on MNIST and 12.1\% versus 6.4\% on Fashion-MNIST.

\begin{figure}[t]
\centering\includegraphics[width=0.93\linewidth]{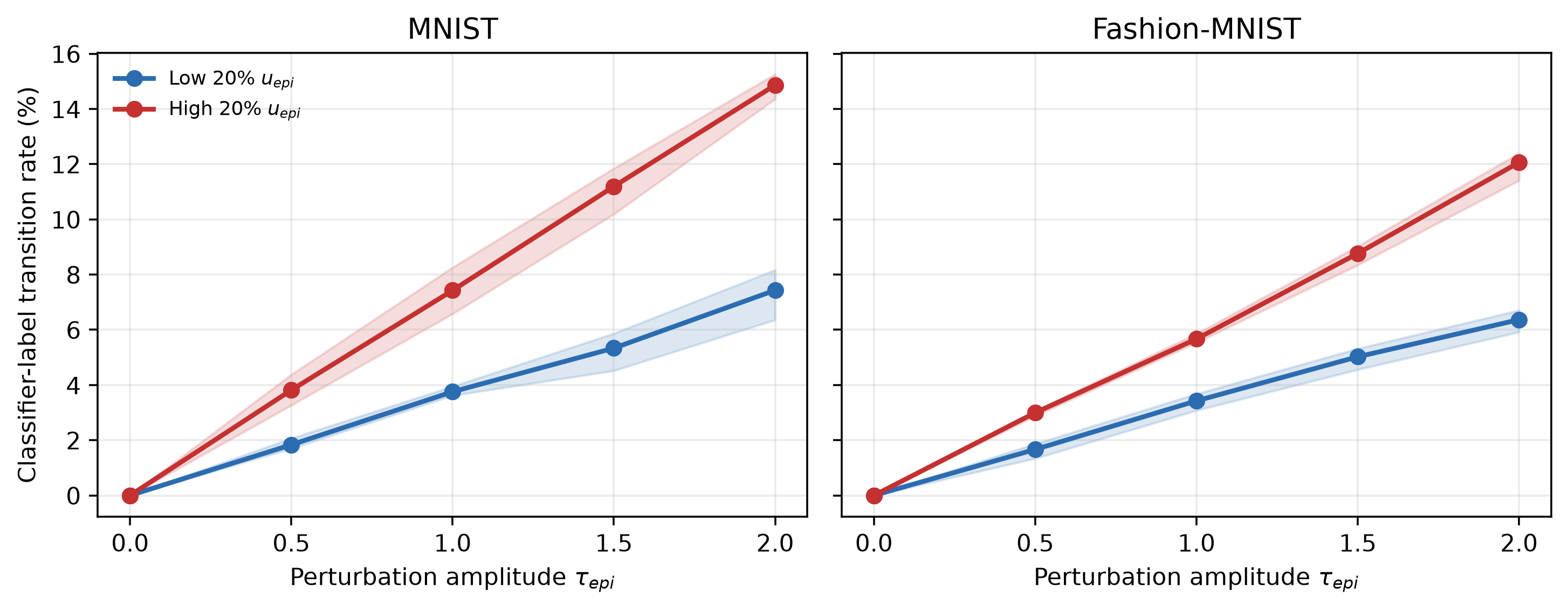}
\caption{Classifier-label transition rate versus perturbation amplitude. Lines are means over three complete ELVAE runs and shaded bands span the three seed values. High-$u_{\mathrm{epi}}$ populations show a consistently steeper dose response on both datasets.}
\label{fig:dose}
\end{figure}

\subsection{Dual-role controls and score comparisons}

The left panel of Fig. 4 tests the dual role of $u_{\mathrm{epi}}$. Under ordinary uncertainty scaling, the three-seed mean high/low transition-rate ratios are 1.978 on MNIST and 1.664 on Fashion-MNIST. Fixing perturbation amplitude to the coordinate-wise class median reduces them to 1.326 and 1.156; within-class permutation gives 1.327 and 1.159. Therefore, a substantial portion of the raw separation comes from the designed amplitude difference, but a smaller anchor-dependent sensitivity remains. The residual is consistent across all three MNIST seeds; on Fashion-MNIST it is weaker and one fixed-scale seed is close to null (1.04). The right panel of Fig. 4 compares scores for induced failure in the detailed runs. On MNIST, reconstruction error and $u_{\mathrm{epi}}$ give similar high/low ratios (1.90 and 1.78). On Fashion-MNIST, $u_{\mathrm{epi}}$ is stronger (1.89 versus 1.72). $u_{\mathrm{var}}$, total variance, and $\|\gamma\|$ do not provide a consistent positive stratification. Their departure from unity is not noise: $\|\gamma\|$ stratifies consistently in the reverse direction (induced-failure ratios 0.23 on MNIST and 0.16 on Fashion-MNIST), as does $u_{\mathrm{var}}$ on Fashion-MNIST (0.21). Anchors with large latent norm or large expected conditional variance are systematically less sensitive to the perturbation applied here. The corresponding AUROCs for $u_{\mathrm{epi}}$ are modest: 0.572 on MNIST and 0.579 on Fashion-MNIST. ELVAE therefore supports population stratification, not accurate prediction of individual failures.

\begin{figure}[t]
\centering\includegraphics[width=0.93\linewidth]{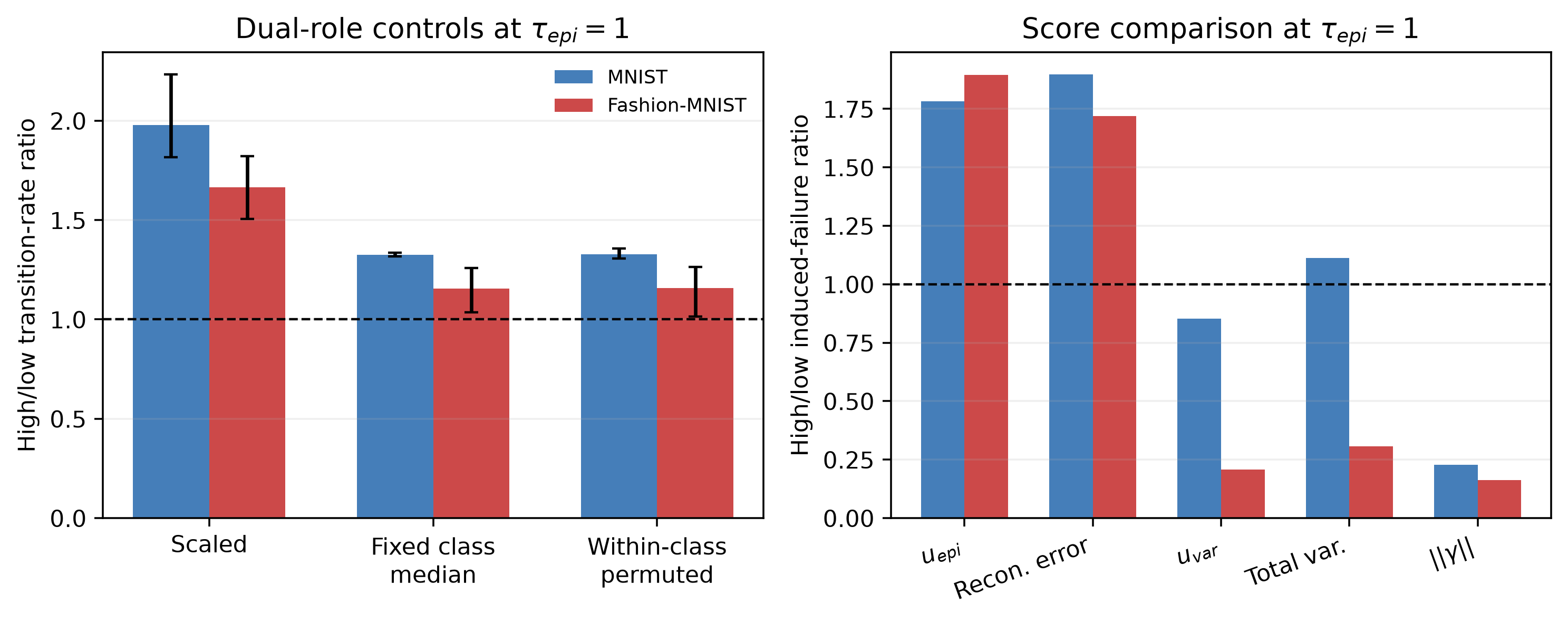}
\caption{Left: high/low semantic-transition ratios at $\tau_{\mathrm{epi}}=1$ under ordinary uncertainty scaling, fixed class-median scaling, and within-class permutation of scale vectors. Bars show three-seed means and error bars span the seed range. Right: detailed-run high/low induced-failure ratios for five anchor scores. The dashed line denotes no separation.}
\label{fig:controls}
\end{figure}

\subsection{Component geometry and the inverse-evidence mechanism}

The focused MNIST experiment tests whether the two terms in Eq. (6) have distinct operational behavior. Without normalization, the mean expected squared displacement at $\tau=1$ is 0.425 for $u_{\mathrm{epi}}$ but 3.735 for $u_{\mathrm{var}}$, an 8.79-fold difference. The corresponding overall transition rates are 4.33\% and 14.10\%. This raw comparison is therefore dominated by perturbation energy and is not evidence that one component is intrinsically more disruptive. After population-level matching by Eq. (26), the overall transition rates

become similar across the full dose response (Fig. 5, left): at $\tau=1$ they are 4.33\% for the $u_{\mathrm{epi}}$ field and 4.61\% for the $u_{\mathrm{var}}$ field. The unchanged 4.33\% for $u_{\mathrm{epi}}$ is required by construction because its population rescaling factor is one. The small paired difference is -0.29 percentage points (95\% anchor-bootstrap interval [-0.34, -0.24]). Thus equal latent energy, rather than the component name, primarily determines the population-average transition rate. The ranking result is different. Table 2 uses the stricter Eq. (27), which gives every anchor identical expected squared displacement. Ranking by $u_{\mathrm{epi}}$ separates transition sensitivity under either learned perturbation field: the high/low ratios are 1.40 under the $u_{\mathrm{epi}}$ field and 1.38 under the $u_{\mathrm{var}}$ field. Ranking by $u_{\mathrm{var}}$ reverses the ordering, giving 0.75 under either field. Equation (7) explains the apparent tension between these opposite directions and the positive scalar correlation $r(u_{\mathrm{epi}},u_{\mathrm{var}})=0.797$. The shared variance scale can make $u_{\mathrm{epi}}$ and $u_{\mathrm{var}}$ move together, while their coordinate-wise ratio isolates $1/\nu$. Ranking by mean inverse evidence strengthens the ratios to 1.69 and 1.65. It correlates more strongly with $u_{\mathrm{epi}}$ ($r=0.925$) than with $u_{\mathrm{var}}$ ($r=0.549$). The two learned fields are not independent geometries. Within an anchor, the median coordinate-wise coefficient of variation of $\nu_k$ is 0.227 (10th--90th percentiles 0.205--0.266), while the median cosine similarity between their anchor-normalized perturbation-scale vectors is 0.987. We therefore use the

\begin{figure}[t]
\centering\includegraphics[width=0.93\linewidth]{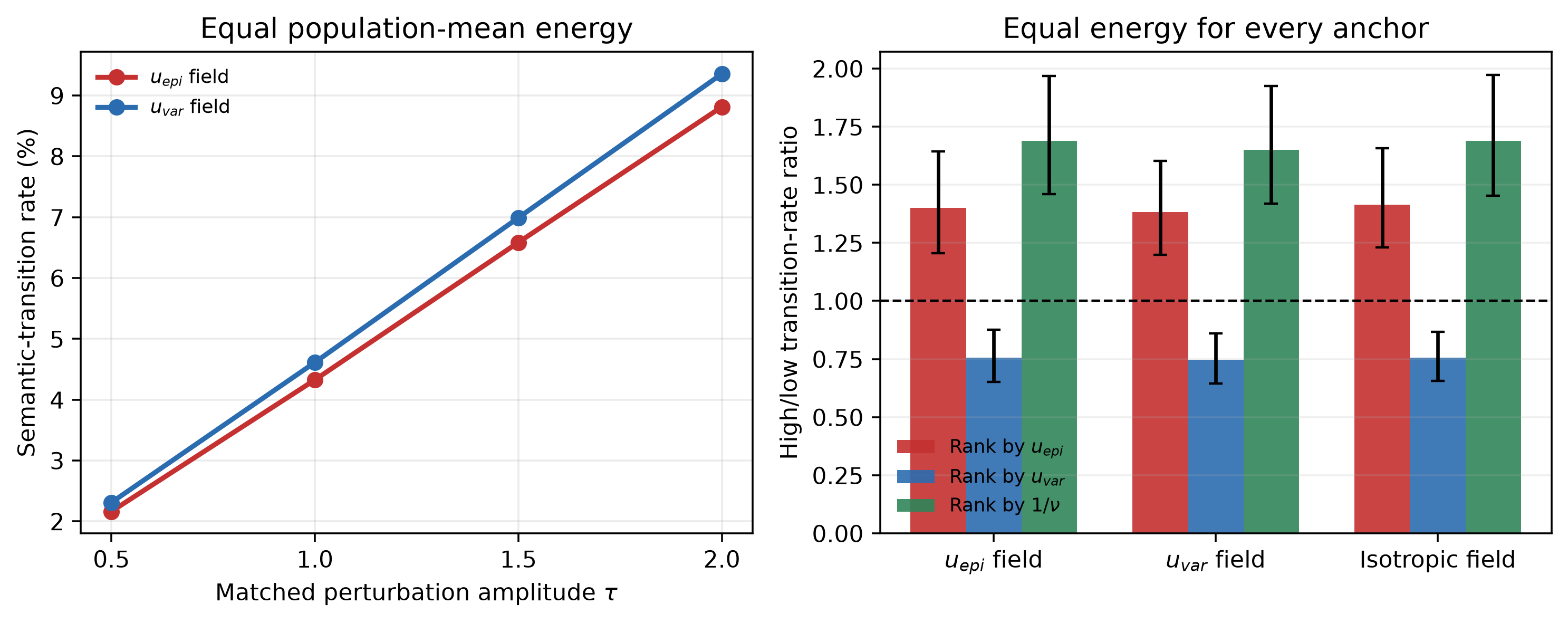}
\caption{MNIST component comparison using 20 paired draws per anchor. Left: overall semantic-transition dose response after population-mean energy matching. Right: high/low ratios after matching energy separately for every anchor, ranked by $u_{\mathrm{epi}}$, $u_{\mathrm{var}}$, or inverse evidence $1/\nu$. The isotropic field has no learned coordinate geometry. Error bars are 95\% anchor-bootstrap intervals; the dashed line denotes no separation.}
\label{fig:component}
\end{figure}

The isotropic field in Eq. (28) is the decisive decoupling control. It contains no learned coordinate allocation, yet inverse evidence retains a high/low ratio of 1.69 ([1.45, 1.97]); $u_{\mathrm{epi}}$ gives 1.42 ([1.23, 1.66]), and $u_{\mathrm{var}}$ again reverses the ordering at 0.76 ([0.66, 0.87]).\footnote{The equality, after rounding, between the 4.33\% low-group entry in the first row of Table 2 and the earlier 4.33\% population rate is coincidental; the former is 4.3275\%.} Thus inverse evidence is an anchor-level sensitivity score rather than an artifact of coupling the score to a learned field. $\nu$ principally governs the sensitivity ordering, and $u_{\mathrm{epi}}$ is a scale-bearing manifestation of that ordering. $u_{\mathrm{var}}$ remains expected conditional latent variability and should not be equated with semantic instability.

The stronger CNN proxy preserves the central ordering (Appendix C). Under the anchor-matched $u_{\mathrm{epi}}$ and $u_{\mathrm{var}}$ fields, $1/\nu$ gives ratios 1.68 and 1.68, while $u_{\mathrm{epi}}$ gives 1.49 and 1.50; all four bootstrap intervals exclude one. By contrast, $u_{\mathrm{var}}$ gives 1.00 under both fields. Because this check uses five rather than 20 draws per anchor, it validates the ordering direction but is not a direct effect-size comparison. Thus the inverse-evidence mechanism is not an artifact of the original 94.79\%-accurate MLP classifier.

\begin{table}[t]
\centering\small
\caption{MNIST semantic transitions at $\tau=1$ with equal expected perturbation energy for every anchor. Each rate averages 20 paired draws per anchor. Intervals are 95\% anchor-bootstrap intervals for the high/low ratio.}
\resizebox{\linewidth}{!}{%
\begin{tabular}{llrrrr}
\toprule
Perturbation field & Ranking score & Low 20\% & High 20\% & High/low & 95\% interval\\
\midrule
$u_{\mathrm{epi}}$ & $u_{\mathrm{epi}}$ & 4.33\% & 6.06\% & 1.40 & [1.20, 1.63]\\
$u_{\mathrm{epi}}$ & $u_{\mathrm{var}}$ & 6.21\% & 4.69\% & 0.75 & [0.65, 0.87]\\
$u_{\mathrm{epi}}$ & $1/\nu$ & 3.87\% & 6.54\% & 1.69 & [1.44, 1.97]\\
$u_{\mathrm{var}}$ & $u_{\mathrm{epi}}$ & 4.49\% & 6.20\% & 1.38 & [1.19, 1.60]\\
$u_{\mathrm{var}}$ & $u_{\mathrm{var}}$ & 6.46\% & 4.83\% & 0.75 & [0.65, 0.86]\\
$u_{\mathrm{var}}$ & $1/\nu$ & 4.06\% & 6.70\% & 1.65 & [1.42, 1.92]\\
Isotropic & $u_{\mathrm{epi}}$ & 4.52\% & 6.40\% & 1.42 & [1.23, 1.66]\\
Isotropic & $u_{\mathrm{var}}$ & 6.57\% & 4.97\% & 0.76 & [0.66, 0.87]\\
Isotropic & $1/\nu$ & 4.09\% & 6.90\% & 1.69 & [1.45, 1.97]\\
\bottomrule
\end{tabular}%
}
\label{tab:components}
\end{table}

\subsection{Quotient geometry and the trained sensitivity score}

The companion theory sharpens the interpretation of the geometric controls without changing the reported perturbation outcomes. Under the present prior,

\begin{equation}
T=\frac{\alpha-1}{\alpha}\frac{\operatorname{Var}(z\mid y)}{\gamma^2+2}. \tag{29}
\end{equation}

The archived component arrays retain $u_{\mathrm{epi},k}$, $u_{\mathrm{var},k}$, and $\gamma_k$, but not $\alpha_k$ and $c_k$ separately. They therefore determine trained $1/\nu_k=u_{\mathrm{epi},k}/u_{\mathrm{var},k}$ and $\operatorname{Var}(z_k\mid y)$ exactly, but they do not determine the exact coordinate-wise canonical score $T_k$ or 1/$\nu_{\mathrm{can},k}$. We accordingly avoid treating the earlier constant-factor stationary approximation as an exact canonical test. What the available arrays do establish is that the trained inverse-evidence effect is not explained by a simple latent-radius score. Ranking anchors by inverse normalized radius $1/\rho$, with $\rho_k=\gamma_k^2/\operatorname{Var}(z_k\mid y)$, gives high/low transition ratios of 1.39 under each of the $u_{\mathrm{epi}}$, $u_{\mathrm{var}}$, and isotropic fields (Table 4), below the 1.69, 1.65, and 1.69 obtained from trained $1/\nu$. The scalar scores are nearly unrelated (Pearson -0.005; rank correlation 0.130). Equation (29) explains why this is a limited negative control rather than a substitute for $T$: the canonical quotient coordinate contains both the prior radius floor 2 and the factor $(\alpha-1)/\alpha$. The empirical and theoretical claims are therefore deliberately separated. The trained four-output model demonstrates that $1/\nu$ is an effective operational sensitivity-ranking score. The quotient analysis shows that the exact canonical representative contains no fourth reconstruction-visible information coordinate. A discrepancy between trained and canonical allocation, if present, should be attributed to finite parameterization, stochastic optimization, or related amortization effects rather than to reconstruction evidence along the fiber.

\subsection{Compact empirical closure of the quotient prediction}

The independent digits ablation permits all three low-cost checks suggested by the quotient theory. First, none of the 3,600 held-out coordinate values from the trained four-output model violates the canonical ceiling: the maximum trained $1/\nu_k$ is 1.139, the 99.9th percentile is 1.107, and the fraction above $3+\sqrt{10}=6.162$ is 0/3600. This does not prove that the amortized model lies exactly on the canonical section, but it rules out the simplest ceiling-based falsification in this run. Second, the exact canonical score can be computed because $(\gamma,\alpha,\beta,\nu)$ were saved. Across the 450 held-out anchors, Spearman correlation between $S_{\mathrm{train}}=K^{-1}\sum_k 1/\nu_k$ and $S_{\mathrm{can}}=K^{-1}\sum_k 1/\nu_{\mathrm{can},k}$ is 0.912. The correlation between $S_{\mathrm{can}}$ and the anchor mean of $T_k$ is 0.999; the slight departure from one is expected because $S_{\mathrm{can}}$ averages a nonlinear transfer $g(T_k)$ rather than $T_k$ itself. Under the energy-matched fields, trained and canonical scores give similar high/low semantic-transition ratios (Table 3). Thus, in this compact run, most of the trained inverse-evidence ordering is already explained by the prior-calibrated quotient geometry, although a smaller off-section component remains possible. Third, the three-output canonical model reaches reconstruction MSE 0.03387 and mean NIG KL 0.717, compared with 0.03243 and 0.747 for the four-output model. With the same objective weight, the corresponding mean objective values are 0.04822 and 0.04737, a 1.8\% difference. The zero-displacement classifier error is 25.3\% versus 20.0\%, so the simpler network is not uniformly better in this single run. Nevertheless, its canonical sensitivity ratios remain positive and comparable to the four-output model. The experiment therefore gives no evidence that a fourth amortized output is required to obtain the sensitivity ordering, while leaving open a finite-optimization advantage for the four-output parameterization.

\begin{table}[t]
\centering\small
\caption{Compact canonical-closure ablation on the independent scikit-learn digits test set (450 anchors, seed 20260809, 20 draws per anchor, $\tau=1$). Brackets are 95\% anchor-bootstrap intervals. The first two rows use the same trained four-output model but different ranking scores; the third row uses the canonical three-output model.}
\resizebox{\linewidth}{!}{%
\begin{tabular}{llccc}
\toprule
Model & Ranking score & $u_{\mathrm{epi}}$ field & $u_{\mathrm{var}}$ field & Isotropic field\\
\midrule
Four-output trained & $S_{\mathrm{train}}$ & 1.52 [1.05, 2.28] & 1.46 [1.11, 1.95] & 1.52 [1.12, 2.08]\\
Four-output canonical & $S_{\mathrm{can}}$ & 1.51 [1.05, 2.23] & 1.41 [1.08, 1.83] & 1.46 [1.10, 1.94]\\
Three-output canonical & $S_{\mathrm{can}}$ & 1.57 [1.13, 2.19] & 1.36 [1.07, 1.76] & 1.47 [1.13, 1.93]\\
\bottomrule
\end{tabular}%
}
\label{tab:closure}
\end{table}
\begin{table}[t]
\centering\small
\caption{Alternative ranking scores under the per-anchor energy-matched protocol at $\tau=1$, computed from the same 20 paired draws as Table 2. Brackets give 95\% anchor-bootstrap intervals. $\rho_k=\gamma_k^2/\operatorname{Var}(z_k\mid y)$ is a normalized-radius surrogate and is not the canonical quotient score $T$ of Eq. (20).}
\resizebox{\linewidth}{!}{%
\begin{tabular}{llrrrr}
\toprule
Perturbation field & Ranking score & Low 20\% & High 20\% & High/low & 95\% interval\\
\midrule
$u_{\mathrm{epi}}$ & $1/\rho$ & 3.50\% & 4.88\% & 1.39 & [1.18, 1.66]\\
$u_{\mathrm{var}}$ & $1/\rho$ & 3.66\% & 5.07\% & 1.39 & [1.18, 1.64]\\
Isotropic & $1/\rho$ & 3.71\% & 5.17\% & 1.39 & [1.19, 1.65]\\
$u_{\mathrm{epi}}$ & $\|\gamma\|$ & -- & -- & 0.32 & --\\
$u_{\mathrm{var}}$ & $\|\gamma\|$ & -- & -- & 0.32 & --\\
Isotropic & $\|\gamma\|$ & -- & -- & 0.32 & --\\
\bottomrule
\end{tabular}%
}
\label{tab:radius}
\end{table}

\subsection{Anchor reliability does not generalize}

Table 5 addresses the distinct question of zero-displacement anchor reliability. On MNIST, $u_{\mathrm{epi}}$ gives a high/low error ratio of 1.29 and AUROC 0.535. 

On Fashion-MNIST the direction reverses: the ratio is 0.81 and AUROC 0.468. Reconstruction error is substantially stronger and consistent, with ratios 2.83 and 2.43. The Gaussian VAE posterior variance also ranks zero-displacement failure more reliably than ELVAE $u_{\mathrm{epi}}$ in these runs. Two further comparisons should be stated plainly. The matched Gaussian VAE attains a slightly lower zero-displacement classifier error than ELVAE (20.97\% versus 22.04\% on MNIST and 21.62\% versus 22.50\% on Fashion-MNIST), and its reconstruction error is at least as strong an anchor score as ELVAE's (Table 5). ELVAE is therefore not offered as an improvement in generation quality or in baseline failure detection. Consequently, the perturbation-sensitivity result must not be presented as evidence that $u_{\mathrm{epi}}$ is a universal quality score.

\begin{table}[t]
\centering\small
\caption{Zero-displacement anchor reliability in the detailed runs.}
\resizebox{\linewidth}{!}{%
\begin{tabular}{llrrrr}
\toprule
Dataset & Score & Low error & High error & Ratio & AUROC\\
\midrule
MNIST & ELVAE $u_{\mathrm{epi}}$ & 22.40\% & 28.80\% & 1.29 & 0.535\\
& ELVAE reconstruction error & 12.60\% & 35.65\% & 2.83 & 0.630\\
& VAE posterior variance & 18.70\% & 29.90\% & 1.60 & 0.563\\
& VAE reconstruction error & 11.75\% & 35.80\% & 3.05 & 0.645\\
\addlinespace
Fashion-MNIST & ELVAE $u_{\mathrm{epi}}$ & 30.10\% & 24.40\% & 0.81 & 0.468\\
& ELVAE reconstruction error & 14.15\% & 34.35\% & 2.43 & 0.618\\
& VAE posterior variance & 22.05\% & 26.60\% & 1.21 & 0.525\\
& VAE reconstruction error & 13.95\% & 34.55\% & 2.48 & 0.617\\
\bottomrule
\end{tabular}%
}
\label{tab:anchor}
\end{table}

\subsection{Seed-to-seed reproducibility}

Table 6 reports all six ELVAE runs. The high/low semantic-transition ratio is greater than one in every run. It ranges from 1.82 to 2.23 on MNIST and from 1.50 to 1.82 on Fashion-MNIST. The induced-failure ratio ranges from 1.78 to 2.26 and from 1.58 to 2.36. By contrast, the zero-displacement anchor-reliability ratio remains above one for MNIST but below one for Fashion-MNIST. This pattern motivates the paper's narrower conclusion.

\begin{table}[t]
\centering\small
\caption{Complete-run reproducibility at $\tau_{\mathrm{epi}}=1$.}
\resizebox{\linewidth}{!}{%
\begin{tabular}{lrrrr}
\toprule
Dataset/seed & Control error & Anchor ratio & Transition ratio & Induced-failure ratio\\
\midrule
MNIST 1 & 23.57\% & 1.06 & 1.88 & 2.10\\
MNIST 2 & 24.50\% & 1.11 & 2.23 & 2.26\\
MNIST 20260809 & 22.04\% & 1.29 & 1.82 & 1.78\\
Fashion-MNIST 1 & 22.30\% & 0.86 & 1.66 & 2.36\\
Fashion-MNIST 2 & 21.90\% & 0.89 & 1.50 & 1.58\\
Fashion-MNIST 20260809 & 22.50\% & 0.81 & 1.82 & 1.89\\
\bottomrule
\end{tabular}%
}
\label{tab:seeds}
\end{table}

\section{Discussion}

The experiments separate three ideas that are easily conflated. First, the NIG hierarchy produces a mathematically well-defined variance of latent location. Second, that quantity can be used as a designed sampling amplitude. Third, its numerical value may or may not predict empirical reliability independently of the sampling rule. Only the first statement follows from the model definition. The second is an operational construction. The third requires data. Across handwritten symbols and apparel objects, the most reproducible empirical property is perturbation sensitivity. High-$u_{\mathrm{epi}}$ populations undergo more classifier-label changes as $\tau_{\mathrm{epi}}$ increases. The effect is not exclusively a consequence of larger perturbations: fixed-scale and permuted-scale controls retain smaller high/low differences. Nevertheless, the reduction under these controls is substantial. Any claim that $u_{\mathrm{epi}}$ reveals an intrinsic property of the anchor must therefore report a scale-matched control. The component experiment strengthens and narrows that interpretation. At equal population-mean energy, $u_{\mathrm{epi}}$ and $u_{\mathrm{var}}$ fields produce nearly the same overall dose response, so neither component is intrinsically much more disruptive at a fixed latent energy. After energy is equalized for every anchor, the stronger and more mechanistic ranking variable is $1/\nu$: it stratifies transition sensitivity under both learned fields more sharply than $u_{\mathrm{epi}}$, whereas $u_{\mathrm{var}}$ ranks it in the opposite direction. Because those two learned fields have highly similar normalized scale vectors, the isotropic control is essential: it removes learned coordinate geometry yet preserves the inverse-evidence ratio of 1.69. The informative result therefore lies in the hierarchy's anchor-level evidence allocation, not in using $\sqrt{u_{\mathrm{epi}}}$ as a larger step size or in coupling a score to its own perturbation field. This separation is operational rather than ontological: classifier transitions establish different empirical roles for the learned quantities, but do not prove that any one of them coincides with a unique real-world source of uncertainty. The compact canonical-closure ablation adds a useful qualification. In that independent run, trained and canonical anchor scores are strongly aligned ($\rho_s$ = 0.912), no coordinate violates the analytic canonical ceiling, and the three-output model retains transition-sensitivity ratios comparable to the four-output model. This favors the parsimonious interpretation that much of the operational ordering can be expressed through prior-calibrated quotient geometry. At the same time, the four-output model has slightly lower reconstruction MSE and lower zero-displacement classifier error in the one-seed ablation, so finite amortized parameterization can still matter. Because the headline 10,000-anchor MNIST archive lacks $\alpha_k$ and $c_k$ separately, the compact check should be viewed as supporting closure rather than a retrospective replacement of the main experiment. The correction analysis changes the interpretation further. Perturbation can move an incorrectly generated anchor back into the intended class. In the detailed Fashion-MNIST run, correction is 8.16\% in the high-$u_{\mathrm{epi}}$ group versus 3.79\% in the low group. Calling all high-$u_{\mathrm{epi}}$ perturbations ``risk'' would ignore this bidirectionality. A more defensible description is semantic instability or transition sensitivity. For stress testing, either direction is informative because it marks proximity to a decision boundary. For synthetic-data augmentation, however, transitions must be screened by the downstream task. The companion theory gives a sharper interpretation [21]. Reconstruction identifies only the quotient $(\gamma,\alpha,c)$ and is exactly invariant along the one-dimensional ($\nu$, $\beta$) fiber; the complete prior and forward KL select the canonical point on that fiber. Canonical inverse allocation is therefore a monotone re-encoding of quotient geometry, not a fourth independent reconstruction-visible coordinate. The empirical finding that trained $1/\nu$ is the strongest sensitivity-ranking score is instead an operational property of the finite four-output amortized model. This distinction is substantive rather than semantic: it prevents the sensitivity result from being misread as evidence that the reconstruction likelihood learned an additional degree of freedom. The radius comparison is correspondingly a negative control, not an identification result. The 1.39 ratio from inverse normalized radius shows that trained $1/\nu$ is not merely a rescaled latent radius, but the exact canonical score also contains the prior radius floor and the tail-dependent factor in Eq. (29). Because no reconstruction residual enters the canonical fiber selector, neither trained nor canonical inverse allocation is theoretically guaranteed to calibrate baseline reconstruction failure; the Fashion-MNIST reversal in Table 5 is consistent with that separation. Reconstruction error is the strongest baseline for zero-displacement anchor reliability. This is unsurprising: it directly measures how faithfully the particular encoder--decoder pair reproduces an input. $u_{\mathrm{epi}}$ answers a different question about the learned higher-order latent posterior. ELVAE is therefore not justified as a replacement for reconstruction-error screening. Its potential value is as a coordinate-wise control field that can modulate how a generator explores the local latent neighborhood. Several limitations remain. Both datasets are low-resolution and the decoders are intentionally compact; pixel-MSE training produces smoothed images. A frozen classifier is a semantic proxy rather than a human or task-specific validity assessment. The CNN check reduces concern about the original MNIST MLP's 94.79\% accuracy but does not replace human validation. The class-conditional ranking used here does not provide a globally comparable uncertainty threshold. The main two-dataset study uses only five draws per anchor, and the AUROCs remain modest. The 20-draw component comparison is intentionally minimal and uses one trained MNIST checkpoint; replication across model seeds and richer datasets remains necessary. The archived headline MNIST component arrays omit $\alpha$ and $c$ separately, so an exact retrospective trained-versus-canonical comparison on that specific 10,000-anchor run is not reconstructible from the saved arrays alone. Section 5.6 therefore reports an independent compact closure experiment with all raw outputs saved, including the ceiling audit, exact canonical ranking, and a three-output ablation. Repeating that ablation on the full MNIST checkpoint would strengthen the architecture comparison but is not required for the mathematical reduction. Extension to natural RGB images will require a stronger perceptual generator. Modern diffusion or flow decoders conditioned on ELVAE uncertainty are a more appropriate next step than scaling the present compact VAE. Future work should evaluate whether $u_{\mathrm{epi}}$ retains scale-matched sensitivity in stronger generators, compare against deep ensembles and other predictive-uncertainty methods [18, 19, 20], and measure downstream utility. A particularly direct study would train classifiers with low-$u_{\mathrm{epi}}$ synthetic augmentation and separately use high-$u_{\mathrm{epi}}$ perturbations for robustness testing. In medical imaging, task-specific physics constraints and expert review would be required before interpreting uncertainty-stratified samples as clinically meaningful.

\section{Conclusion}

ELVAE separates latent-location uncertainty from conditional variability through an input-dependent NIG hierarchy, but reconstruction itself sees only the three-coordinate quotient $(\gamma,\alpha,c)$. The companion theory shows that the complete prior and forward KL select the canonical fourth-coordinate representative; under the experimental prior, canonical $1/\nu_{\mathrm{can}}$ is a monotone transform of $T=c/[\alpha(\gamma^2+2)]$ rather than an independent information channel [21]. The empirical result is complementary: in the finite four-output model, trained $1/\nu$ is the strongest sensitivity-ranking score after perturbation energy is equalized, with ratios 1.69, 1.65, and 1.69 under the $u_{\mathrm{epi}}$, $u_{\mathrm{var}}$, and isotropic fields. This ordering survives removal of learned perturbation geometry and a higher-accuracy CNN check, while the normalized-radius surrogate reaches only 1.39. At the same time, $u_{\mathrm{epi}}$ is not a universal anchor-quality score and perturbations can induce both failures and corrections. The resulting picture is sharper than the original formulation: reconstruction identifies three latent coordinates, the prior-relative KL selects the canonical allocation, and the trained four-output realization supplies an operational sensitivity score whose usefulness must be evaluated separately from image quality. A compact independent ablation further shows strong trained--canonical rank agreement ($\rho_s$ = 0.912), no violation of the analytic ceiling, and comparable sensitivity ordering from a three-output canonical encoder, suggesting that the fourth output is not structurally required for the observed ordering even though it can affect finite-network optimization.

\section*{Declaration of competing interest}
The author declares no competing financial interest or personal relationship that could have influenced the work reported in this paper.

\section*{Author--AI Collaboration.}
Generative AI tools, ChatGPT and Claude, were used to assist the author in brainstorming, formulation, analysis, drafting, simulation, checking, and polishing. The author conceptualized the idea and frame, supervised the AI models, and takes responsibility for the content.

\appendix
\section{Closed-form NIG divergence}
For $q=\operatorname{NIG}(\gamma,\nu,\alpha,\beta)$ and $p_0=\operatorname{NIG}(\gamma_0,\nu_0,\alpha_0,\beta_0)$ under the inverse-gamma parameterization used in Eq. (3), the coordinate-wise divergence implemented in the experiments is
\begin{align}
D_{\mathrm{KL}}(q\|p_0)={}&\alpha_0\log\frac{\beta}{\beta_0}-\log\Gamma(\alpha)+\log\Gamma(\alpha_0)+(\alpha-\alpha_0)\psi(\alpha)-\alpha+\frac{\alpha\beta_0}{\beta}\nonumber\\
&+\frac12\left[\log\frac{\nu}{\nu_0}+\frac{\nu_0}{\nu}-1+\frac{\nu_0\alpha(\gamma-\gamma_0)^2}{\beta}\right]. \tag{A.1}
\end{align}
where $\psi$ is the digamma function. The first line is the inverse-gamma divergence and the second is the expected conditional-normal divergence.

\section{Interpretation of paired transitions}
Let $C_i$ denote whether anchor $i$ is classified correctly at $z=\gamma$ and let $A_{ij}(\tau_{\mathrm{epi}})$ denote correctness for draw $j$ after perturbation. The reported induced-failure and induced-correction rates are
\begin{equation}
D(\tau_{\mathrm{epi}})=\Pr(A_{ij}=0\mid C_i=1), \tag{B.1}
\end{equation}
\begin{equation}
R(\tau_{\mathrm{epi}})=\Pr(A_{ij}=1\mid C_i=0). \tag{B.2}
\end{equation}
Their denominators differ, so $D-R$ is not the net error change. The correct identity is
\begin{equation}
\Pr(A=0)-\Pr(C=0)=\Pr(C=1,A=0)-\Pr(C=0,A=1). \tag{B.3}
\end{equation}
The semantic-transition statistic instead compares classifier labels directly,
\begin{equation}
T(\tau_{\mathrm{epi}})=\Pr(\widehat c_A\ne\widehat c_C), \tag{B.4}
\end{equation}
and therefore captures bidirectional and wrong-to-different-wrong changes.

\section{CNN semantic-proxy robustness}
The convolutional classifier is trained only on the 60,000 real MNIST training images and reaches 98.84\% accuracy on the 10,000 real test images. It is then frozen. Table C.7 repeats the per-anchor energy-matched $\tau=1$ comparison using five common perturbation draws for every anchor. The inverse-evidence and $u_{\mathrm{epi}}$ rankings remain positive under either field, whereas the $u_{\mathrm{var}}$ ranking is neutral.
\begin{table}[h]
\centering\small
\caption{CNN robustness check for MNIST semantic transitions. Brackets give 95\% anchor-bootstrap intervals.}
\resizebox{\linewidth}{!}{%
\begin{tabular}{llrrrr}
\toprule
Perturbation field & Ranking score & Low 20\% & High 20\% & High/low & 95\% interval\\
\midrule
$u_{\mathrm{epi}}$ & $u_{\mathrm{epi}}$ & 4.72\% & 7.03\% & 1.49 & [1.26, 1.77]\\
$u_{\mathrm{epi}}$ & $u_{\mathrm{var}}$ & 5.78\% & 5.80\% & 1.00 & [0.85, 1.19]\\
$u_{\mathrm{epi}}$ & $1/\nu$ & 4.43\% & 7.45\% & 1.68 & [1.42, 2.00]\\
$u_{\mathrm{var}}$ & $u_{\mathrm{epi}}$ & 4.86\% & 7.31\% & 1.50 & [1.28, 1.78]\\
$u_{\mathrm{var}}$ & $u_{\mathrm{var}}$ & 6.10\% & 6.10\% & 1.00 & [0.85, 1.17]\\
$u_{\mathrm{var}}$ & $1/\nu$ & 4.56\% & 7.68\% & 1.68 & [1.42, 1.99]\\
\bottomrule
\end{tabular}%
}
\label{tab:cnn}
\end{table}

\section{Canonical selector and monotone quotient score}
For the general-prior derivation, exact three-coordinate reduction, transfer function, amplitude bound, and residual-gauge analysis, see the companion theoretical study [21]. We retain the experimental-prior derivation here until the empirical closure analyses are complete. Fix $(\gamma,\alpha,c)$ and the experimental prior $p_0=\operatorname{NIG}(0,1,3,1)$, and restrict the NIG divergence in Eq. (A.1) to
\[
\beta=\frac{c\nu}{1+\nu}.
\]
After dropping terms independent of $\nu$, the fiber objective is
\begin{equation}
f(\nu)=\frac72\log\nu-3\log(1+\nu)+\frac{B}{\nu}+C,\qquad B=\frac12+\frac{\alpha}{c}\left(1+\frac{\gamma^2}{2}\right), \tag{D.1}
\end{equation}
where $C$ is independent of $\nu$. Differentiating gives
\begin{equation}
f'(\nu)=\frac{7/2}{\nu}-\frac{3}{1+\nu}-\frac{B}{\nu^2}. \tag{D.2}
\end{equation}
Multiplication by $2\nu^2(1+\nu)>0$ yields
\begin{equation}
\nu^2+(7-2B)\nu-2B=0. \tag{D.3}
\end{equation}
The roots have product $-2B<0$, so exactly one root is positive, namely Eq. (18). Moreover, $f(\nu)\to\infty$ as $\nu\downarrow0$ and as $\nu\to\infty$, hence this positive stationary point is the unique global minimum.

The no-loss reduction stated in Section 2.4 follows immediately. At fixed $(\gamma,\alpha,c)$, the reconstruction term is constant along the fiber, so minimizing the full objective over $\nu$ is exactly the same as replacing the KL by its fiber minimum $R_{\mathrm{can}}(\gamma,\alpha,c)$ and then optimizing the three quotient coordinates.

For the prior $\operatorname{NIG}(0,1,3,1)$, define
\[
T=\frac{c}{\alpha(\gamma^2+2)}.
\]
Then $B=\tfrac12(1+1/T)$, which is strictly decreasing in $T$. Differentiating Eq. (18) with respect to $B$ gives a strictly positive derivative, so $\nu_{\mathrm{can}}$ increases with $B$ and $1/\nu_{\mathrm{can}}$ decreases with $B$. Therefore $1/\nu_{\mathrm{can}}=g(T)$ is strictly increasing in $T$, proving the rank equivalence stated in Section 2.4.

For $\alpha>1$, $\operatorname{Var}(z\mid y)=c/(\alpha-1)$, so
\[
T=\frac{\alpha-1}{\alpha}\frac{\operatorname{Var}(z\mid y)}{\gamma^2+2}.
\]
This identity also shows why the normalized-radius statistic $\operatorname{Var}(z\mid y)/\gamma^2$ used in Table 4 is only a surrogate: it omits both the prior radius floor and the tail-dependent factor.


\begin{thebibliography}{22}
\bibitem{r1} D. P. Kingma, M. Welling, Auto-encoding variational bayes, in: International Conference on Learning Representations, 2014.
\bibitem{r2} D. P. Kingma, M. Welling, An introduction to variational autoencoders, Foundations and Trends in Machine Learning 12 (4) (2019) 307--392.
\bibitem{r3} M. Sensoy, L. Kaplan, M. Kandemir, Evidential deep learning to quantify classification uncertainty, in: Advances in Neural Information Processing Systems, Vol. 31, 2018.
\bibitem{r4} A. Amini, W. Schwarting, A. Soleimany, D. Rus, Deep evidential regression, in: Advances in Neural Information Processing Systems, Vol. 33, 2020.
\bibitem{r5} B. Charpentier, D. Z\"ugner, S. G\"unnemann, Posterior network: Uncertainty estimation without OOD samples via density-based pseudo-counts, in: Advances in Neural Information Processing Systems, Vol. 33, 2020, pp. 1356--1367.
\bibitem{r6} B. Charpentier, O. Borchert, D. Z\"ugner, S. Geisler, S. G\"unnemann, Natural posterior network: Deep bayesian predictive uncertainty for exponential family distributions, in: International Conference on Learning Representations, 2022.
\bibitem{r7} D. Ulmer, C. Hardmeier, J. Frellsen, Prior and posterior networks: A survey on evidential deep learning methods for uncertainty estimation, Transactions on Machine Learning Research (2023).
\bibitem{r8} V. Bengs, E. H\"ullermeier, W. Waegeman, Pitfalls of epistemic uncertainty quantification through loss minimisation, in: Advances in Neural Information Processing Systems, Vol. 35, 2022.
\bibitem{r9} N. Meinert, J. Gawlikowski, A. Lavin, The unreasonable effectiveness of deep evidential regression, in: Proceedings of the AAAI Conference on Artificial Intelligence, Vol. 37, 2023, pp. 9134--9142.
\bibitem{r10} M. Itkina, B. Ivanovic, R. Senanayake, M. J. Kochenderfer, M. Pavone, Evidential sparsification of multimodal latent spaces in conditional variational autoencoders, in: Advances in Neural Information Processing Systems, Vol. 33, 2020.
\bibitem{r11} G. Baykal, M. Kandemir, G. Unal, EdVAE: Mitigating codebook collapse with evidential discrete variational autoencoders, Pattern Recognition 156 (2024) 110792.
\bibitem{r12} H. Takahashi, T. Iwata, Y. Yamanaka, M. Yamada, S. Yagi, Student-$t$ variational autoencoder for robust density estimation, in: Proceedings of the 27th International Joint Conference on Artificial Intelligence (IJCAI), 2018, pp. 2696--2702.
\bibitem{r13} N. Abiri, M. Ohlsson, Variational auto-encoders with student's $t$-prior, in: Proceedings of the 27th European Symposium on Artificial Neural Networks, Computational Intelligence and Machine Learning (ESANN), 2019.
\bibitem{r14} E. Mathieu, T. Rainforth, N. Siddharth, Y. W. Teh, Disentangling disentanglement in variational autoencoders, in: Proceedings of the 36th International Conference on Machine Learning (ICML), Vol. 97, PMLR, 2019, pp. 4402--4412.
\bibitem{r15} J. Kim, J. Kwon, M. Cho, H. Lee, J.-H. Won, $t^3$-variational autoencoder: Learning heavy-tailed data with student's $t$ and power divergence, in: International Conference on Learning Representations (ICLR), 2024.
\bibitem{r16} H. Xiao, K. Rasul, R. Vollgraf, Fashion-MNIST: A novel image dataset for benchmarking machine learning algorithms, arXiv preprint arXiv:1708.07747 (2017).
\bibitem{r17} Y. LeCun, L. Bottou, Y. Bengio, P. Haffner, Gradient-based learning applied to document recognition, Proceedings of the IEEE 86 (11) (1998) 2278--2324.
\bibitem{r18} A. Kendall, Y. Gal, What uncertainties do we need in bayesian deep learning for computer vision?, in: Advances in Neural Information Processing Systems, Vol. 30, 2017.
\bibitem{r19} B. Lakshminarayanan, A. Pritzel, C. Blundell, Simple and scalable predictive uncertainty estimation using deep ensembles, in: Advances in Neural Information Processing Systems, Vol. 30, 2017.
\bibitem{r20} Y. Ovadia, E. Fertig, J. Ren, Z. Nado, D. Sculley, S. Nowozin, J. Dillon, B. Lakshminarayanan, J. Snoek, Can you trust your model's uncertainty? evaluating predictive uncertainty under dataset shift, in: Advances in Neural Information Processing Systems, Vol. 32, 2019.
\bibitem{r21} G. Wang, Theoretical study on the evidential learning-based variational autoencoder, arXiv preprint, submitted (2026).
\bibitem{r22} F. Pedregosa, G. Varoquaux, A. Gramfort, V. Michel, B. Thirion, O. Grisel, M. Blondel, P. Prettenhofer, R. Weiss, V. Dubourg, J. VanderPlas, A. Passos, D. Cournapeau, M. Brucher, M. Perrot, E. Duchesnay, Scikit-learn: Machine learning in Python, Journal of Machine Learning Research 12 (2011) 2825--2830.
\end{thebibliography}
\end{document}